\documentclass[journal]{IEEEtran}

\ifCLASSINFOpdf
\else
   \usepackage[dvips]{graphicx}
\fi
\usepackage{url}

\usepackage{graphicx}

\usepackage{tabularx}
\usepackage{ragged2e} 
\usepackage{booktabs}
\usepackage{amsmath}
\usepackage{multirow}
\usepackage{balance}
\usepackage[colorlinks,citecolor=blue,linkcolor=blue]{hyperref}

\begin{document}

\title{D-Aug: Enhancing Data Augmentation for Dynamic LiDAR Scenes}

\author{Jiaxing Zhao, Peng Zheng and Rui Ma
\thanks{Corresponding authors: Peng Zheng and Rui Ma.}
\thanks{Jiaxing Zhao, Peng Zheng, and Rui Ma are with the School of Artificial Intelligence, Jilin University, Changchun 130000, China (email: zhaojx9921@mails.jlu.edu.cn; zhengpeng22@mails.jlu.edu.cn; ruim@jlu.edu.cn)}}

\maketitle

\begin{abstract}
Creating large LiDAR datasets with pixel-level labeling poses significant challenges. While numerous data augmentation methods have been developed to reduce the reliance on manual labeling, these methods predominantly focus on static scenes and they overlook the importance of data augmentation for dynamic scenes, which is critical for autonomous driving. To address this issue, we propose D-Aug, a LiDAR data augmentation method tailored for augmenting dynamic scenes. D-Aug extracts objects and inserts them into dynamic scenes, considering the continuity of these objects across consecutive frames. For seamless insertion into dynamic scenes, we propose a reference-guided method that involves dynamic collision detection and rotation alignment. Additionally, we present a pixel-level road identification strategy to efficiently determine suitable insertion positions. We validated our method using the nuScenes dataset with various 3D detection and tracking methods. Comparative experiments demonstrate the superiority of D-Aug.

\end{abstract}


\section{Introduction}
\IEEEPARstart{D}UE to its precise range-sensing ability for capturing 3D geometric information, LiDAR sensors have found widespread use in various applications, particularly in the field of Autonomous Driving (AD). Recently, the most promising approach for processing LiDAR data involves training deep neural networks in various downstream applications \cite{Tang_Liu_Zhao_Lin_Lin_Wang_Han_2020, Tu_Wang_Liu_2021, Zhu_Zhou_Wang_Hong_Ma_Li_Li_Lin_2021, Charles_Su_Kaichun_Guibas_2017, Qi_Yi_Su_Guibas_2017, Xu_Wu_Wang_Zhan_Vajda_Keutzer_Tomizuka_2020}. However, this approach requires plenty of labeled data \cite{Huang_Wang_Cheng_Zhou_Geng_Yang_2020}, especially in 3D object detection and tracking tasks \cite{Sadjadpour_Li_Ambrus_Bohg_2022, chen2023focalformer3d, bai2022transfusion, liu2022gnn, qi2018frustum, lang2019pointpillars, zheng2022beyond, pang2022simpletrack}. Unfortunately, the manual collection and labeling of such data are time-consuming and labor-intensive \cite{Fang_Zuo_Zhou_Jin_Wang_Zhang_2021}, impeding comprehensive analysis and understanding of LiDAR point clouds. To alleviate the burden of data labeling, data augmentation \cite{hahner2020quantifying, Simonyan_Zisserman_2015} has emerged as a prevailing method. It aims to effectively reduce the need for data labeling by enriching the training set through transformations of existing data. 

\begin{figure}[t]
  \centering
  \includegraphics[width=1\linewidth]{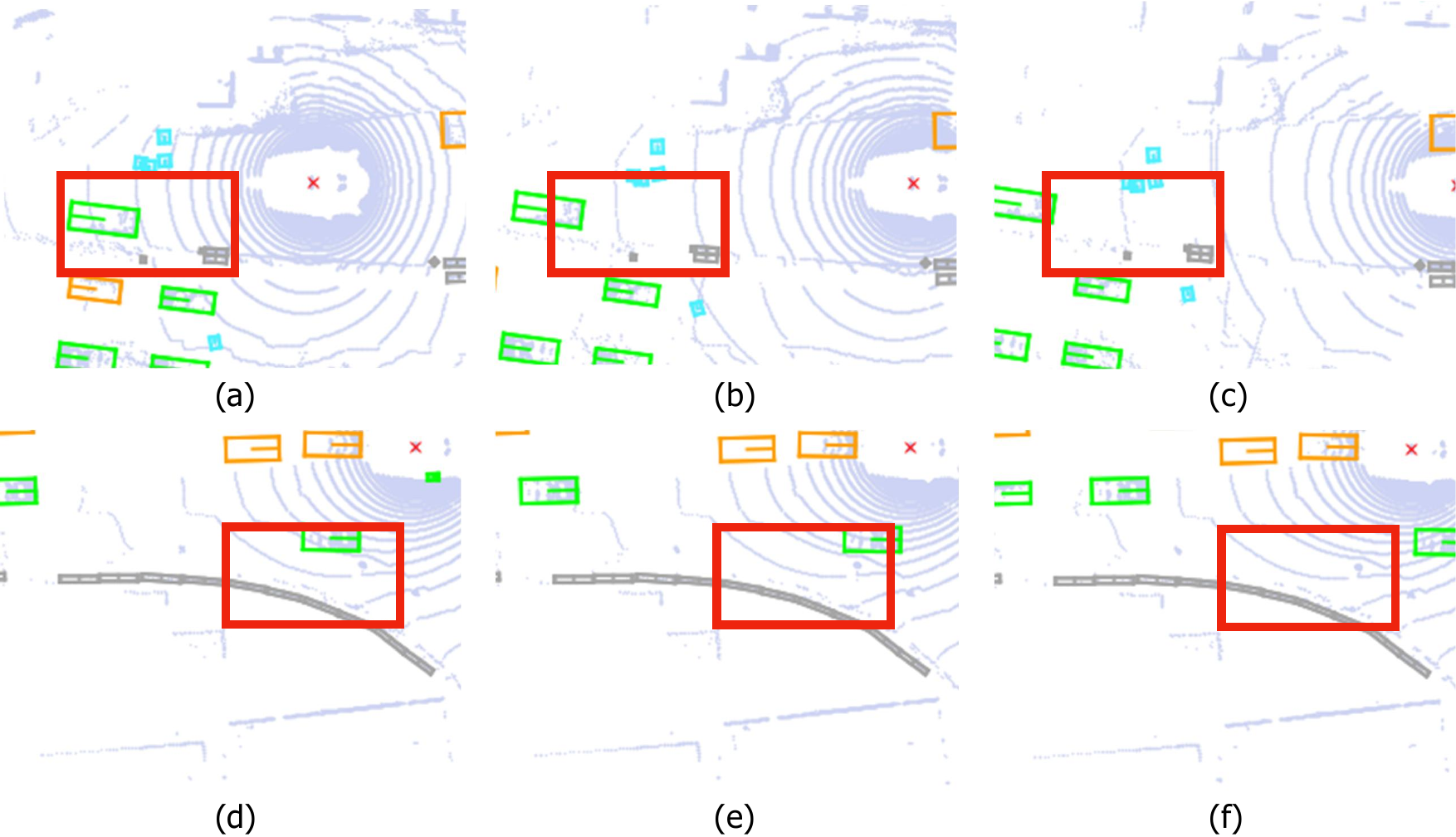}
    \caption{Illustration of augmented LiDAR data. The three figures in the first row as well as those figures in the second row , display the augmented point clouds for three successive frames from two distinct scenes. The orange and green bounding boxes represent the original and inserted objects, respectively. Notably, the areas within the red rectangles emphasize the relative movement between the inserted objects and the stationary obstacles (grey boxes).}
  \label{fig: 1}
\end{figure}

Some global data augmentation methods \cite{Yan_Mao_Li_2018, Lang_Vora_Caesar_Zhou_Yang_Beijbom_2019, Shi_Guo_Jiang_Wang_Shi_Wang_Li_2020, Shi_Wang_Li_2019, He_Zeng_Huang_Hua_Zhang_2020, Hahner_Dai_Liniger_Gool_2020, Yang_Shi_Wang_Li_Qi_2021, Chen_Kornblith_Norouzi_Hinton_2020} involve manipulating the entire LiDAR dataset on a global scale, such as random scaling, flipping, and rotation. Conversely, other methods \cite{Yan_Mao_Li_2018, Shi_Wang_Li_2019, vsebek2022real3d, Xiao_Huang_Guan_Cui_Lu_Shao_2022} focus on object-level augmentation. For instance, LiDAR-Aug \cite{Fang_Zuo_Zhou_Jin_Wang_Zhang_2021} integrates objects rendered from Computer-Aided Design (CAD) models into the original LiDAR point clouds. Despite the variety of data augmentation methods, they predominantly focus on statically augmenting the current frame, overlooking the continuity of augmented objects across consecutive frames. Ensuring this continuity is crucial for object detection and tracking tasks, and addressing this aspect is essential to maintain the realism of augmented data.

\begin{figure*}[t]
  \centering
  \includegraphics[width=1\linewidth]{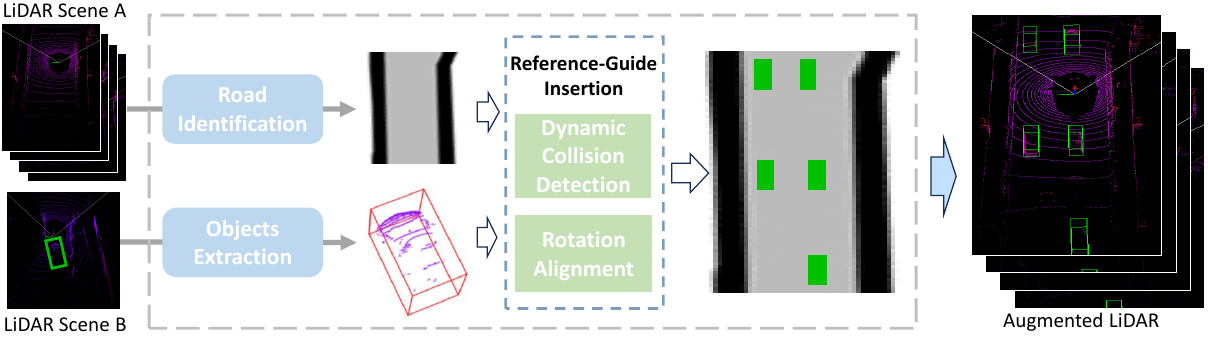}
  \caption{Overview of D-Aug. Consecutive point cloud frames A are processed: available insertion positions are first determined through road identification. Subsequently, objects are extracted from point cloud B by calculating direction vectors. Finally, these extracted objects are inserted into each frame of A using a reference-guided insertion approach, which incorporates dynamic collision detection and rotation alignment. Notably, the inserted objects are rotated to align with the traffic flow in the dynamic scene, resulting in augmented dynamic scenes. }
  \label{fig: 2}
\end{figure*}

To address the aforementioned limitations, we introduce D-Aug, a novel LiDAR data augmentation method tailored for dynamic scenes. Unlike previous approaches, our method focuses on improving the continuity of objects across successive frames. We introduce a pixel-level road identification technique to locate available insertion positions within the scene, ensuring they align with the actual traffic flow. Additionally, we employ a dynamic collision detection algorithm to guarantee that inserted objects remain collision-free in dynamic scenes. In Figure \ref{fig: 1}, we illustrate augmented LiDAR data, where we use stationary obstacles as references to highlight how the augmented objects maintain dynamic continuity in successive frames.

The key contributions of our method are outlined as follows:
\begin{enumerate}
\item We propose D-Aug, a novel LiDAR data augmentation method tailored for dynamic scenes, ensuring the continuity of inserted objects across successive frames.
\item Our proposed method is evaluated on the publicly available nuScenes \cite{nuScenes} dataset, demonstrating significant improvements in 3D object detection and tracking performance compared to various baselines.
\end{enumerate}

\section{Proposed Method}
The following subsections present our approach: first, a pixel-level road identification method is introduced, which proves to be more efficient than the region-based segmentation provided by the nuScenesMap-API. Next, we describe the extraction of objects from a given point cloud. Finally, we discuss the insertion of extracted objects using reference-guided insertion. An overview of the proposed method is illustrated in Fig. \ref{fig: 2}.

\subsection{Pixel-Level Road Identification}

\begin{figure}[t]
  \centering
  \includegraphics[width=1\linewidth]{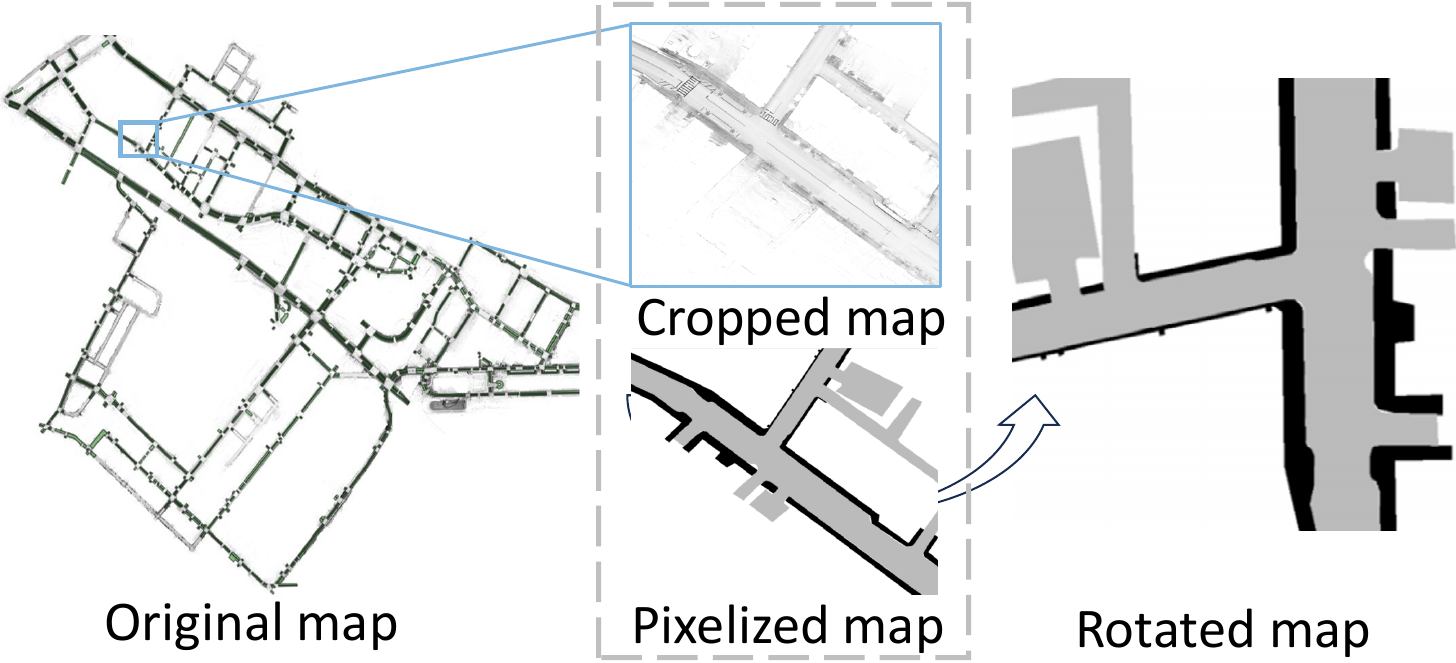}
  \caption{Illustration of pixel-level road identification. The map is cropped, pixelized, and rotated to facilitate road identification based on pixel values. The grey areas represent the roads where the objects can be inserted. }
  \label{fig: 3}
\end{figure}

Before inserting objects, it is essential to find a suitable position to avoid overlap with existing objects in the scene. One possible solution is using nuScenesMap-API, which provides a fine-grained layer-based method to identify roads. However, it tends to be slow and inefficient when dealing with large-scale scenes. To address this issue, we introduce a pixel-level road identification method. Specifically, our method encompasses several key steps, including map cropping, pixelization, and road identification, as shown in Fig. \ref{fig: 3}. 

Processing the entire scene is time-consuming, and LiDAR data far from the vehicle is often irrelevant for most applications. Hence, we crop the map for efficiency. Given LiDAR data, ego-pose, map, and the corresponding map mask, we pixelize the map and convert the position of the vehicle into pixel coordinates on the map. This pixelization is achieved by assigning distinct colors to different classes, where the class of each pixel is provided in the map mask. This approach enables us to determine the class for each pixel through a single query, rather than individual queries for each class in the map mask. Specifically, we assign "grey" to "road". Subsequently, we crop the map with the vehicle's pixel coordinates as the center. The cropped map is then rotated to align with the LiDAR data. Since the map is pixelized and our assignment of "grey" to "road", roads can be directly identified based on the pixel values. During insertion, we validate the inserted position against the map. Specifically, we project the bounding box of inserted objects onto the map. The insertion position is deemed valid only if the projected bounding box does not intersect areas other than the road.

\begin{figure}[t]
  \centering
  \includegraphics[width=1\linewidth]{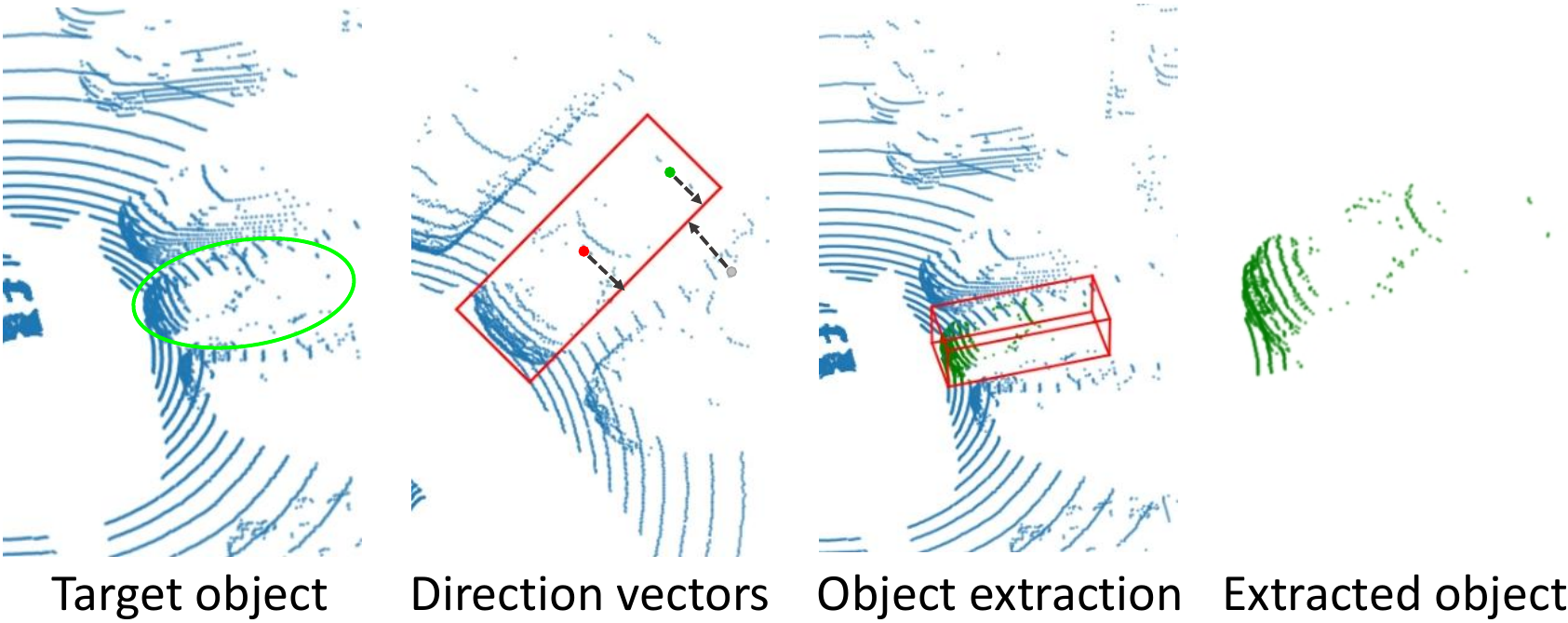}
  \caption{Illustration of object extraction. Points within a bounding box are identified by calculating direction vectors. Points sharing the same direction vectors as the center point of the bounding box are extracted as objects. }
  \label{fig: 4}
\end{figure}

\begin{table*}[t]
\centering
\small 
\caption{Evaluation results of the 3D detection task on the nuScenes validation dataset}
\begin{tabularx}{\textwidth}{@{}l|*{2}{p{0.05\textwidth}}|*{10}{p{0.05\textwidth}}@{}}
\bottomrule
Method    & {mAP} & {NDS}  & {Car} & {Tru} & {Bus} & {Tra} & {C.V.} & {Ped} & {Mot} & {Byc} & {T.C.} & {Bar} \\
\toprule
\bottomrule
CBGS~\cite{zhu2019class} & 50.10 & 61.61       & 81.2  & 51.5    & 66.4  & 37.3       & 16.3   & 77.2  & 38.4   & \textbf{17.7}  & \textbf{57.3}  & \textbf{57.6} \\
 + D-Aug & \textbf{50.68}  & \textbf{61.77}       & \textbf{82.1}  & \textbf{52.8}    & \textbf{67.5}  & \textbf{39.1}       & 16.3   & \textbf{77.7}  & \textbf{40.2}   & 16.4  & \textbf{57.3}  & \textbf{57.5} \\
\hline
PointPillars~\cite{pointpillars} & 43.30  & \textbf{57.50}       & 80.6  & 48.0    & 62.5  & 33.7       & 10.7   & 70.9  & 30.0   & \textbf{4.6}  & 44.0  & 48.0 \\
 + D-Aug & \textbf{43.98}  & 57.46       & \textbf{80.9}  & \textbf{50.1}    & \textbf{63.1}  & \textbf{34.2}       & \textbf{10.8}   & \textbf{71.2}  & \textbf{32.2}   & 3.9 & \textbf{45.2}  & \textbf{48.2} \\
\hline
CenterPoint~\cite{Yin_Zhou_Krahenbuhl_2021} & 59.10  & 66.69       & 85.3  & 57.3    & 71.5  & \textbf{37.9}       & 17.1   & 85.0  & 58.9   & 41.4  & \textbf{69.5}  & 67.1 \\
 + D-Aug & \textbf{59.68}  & \textbf{67.16}       & \textbf{85.5}  & \textbf{58.6}    & \textbf{71.7}  & \textbf{37.9}       & \textbf{18.4}   & \textbf{85.2}  & \textbf{60.0}   & \textbf{42.2}  & 68.9 & \textbf{68.4}  \\
\hline
VoxelNeXt~\cite{Chen_Liu_Zhang_Qi_Jia}  & 60.45  & 66.72       & \textbf{83.9}  & 57.1    & 70.5  & 38.5      & 19.3   & \textbf{84.8}  & \textbf{63.3}   & 49.2  & 69.8 & \textbf{68.0}  \\
 + D-Aug  & \textbf{60.64}  & \textbf{67.07}       & \textbf{83.9}  & \textbf{57.6}    & \textbf{70.7}  & \textbf{39.3}      & \textbf{22.2}   & 84.7  & 62.2   & \textbf{49.8}  & \textbf{70.1} & 65.9  \\
\hline
TransFusion-L~\cite{transfusion} & 63.80  & 68.88       & 86.4  & 52.2    & 72.6  & 44.0       & \textbf{26.6}   & 86.5  & 70.4   & \textbf{57.0}  & \textbf{73.9}  & 68.4 \\
 + D-Aug & \textbf{64.44}  & \textbf{69.11}       & \textbf{86.8}  & \textbf{54.8}    & \textbf{75.0}  & \textbf{45.9}       & 26.1   & \textbf{86.7}  & \textbf{71.4}   & 55.1  & 73.1  & \textbf{69.6} \\
\toprule
\end{tabularx}
\label{label2}
\end{table*}

\subsection{Objects Extraction}

For effective point cloud insertion at the object level, precise object extraction is imperative. This process involves retrieving all point clouds associated with an object along with their corresponding bounding boxes. While the nuScenes \cite{nuScenes} dataset provides bounding box information, extracting the point cloud is intricate due to the mismatch between LiDAR data recorded in the local coordinate system and the bounding boxes recorded in the global coordinate system. By utilizing rotation quaternions, translation vectors, and ego-poses from the calibrated sensor, we transform the point cloud from the local to the global coordinate system, aligning it with the bounding box.
To extract the point cloud within the bounding box, we utilize direction vectors from points to the bounding box faces, as illustrated in Fig. \ref{fig: 4}. A direction vector is defined as a vector starting from the given point and perpendicular to the specified face. Given a bounding box $B$, we compute direction vectors from its center point $c$ to each face $F_i$. For each point $p$ in the point cloud $P$, we calculate its direction vectors in the same manner. Points inside the bounding box share direction vectors with the center point.


\subsection{Reference-Guided Objects Insertion}
Randomly selecting insertion positions can lead to various issues, such as objects floating above the ground or misaligned orientations with the traffic flow, particularly in dynamic scenes. Intuitively, the position of existing objects can guide the insertion. Hence, we introduce a reference-guided insertion algorithm: Initially, an object is randomly chosen as a reference. Next, we search for available insertion positions within a set distance around the reference object. The availability of a position is determined by two factors: ensuring the inserted object remains grounded and avoiding collisions with other objects. If no suitable position is found, other objects are chosen as references for the search algorithm. Upon identifying a suitable insertion position, the extracted objects are first rotated to align with the traffic flow within the scene, facilitating smooth insertion. Finally, the objects are inserted into the target scene, ensuring consistency with its overall layout.

Collision detection in dynamic data augmentation poses more challenges than in static augmentation, as it must account for potential collisions not only in the current frame but also in subsequent frames of the target scene. By considering the velocity vector and position of bounding boxes in the current frame, collision relationships can be computed for each frame. The search for insertion positions continues until no collisions are detected, at which point the searched position is considered an available insertion position.

In a frame $S_i$ from the set $\mathcal{S}=\{ S_0, S_1, \dots, S_K \}$ containing a set of bounding boxes $\mathcal{B}i = \{B_i^1, B_i^2, \dots, B_i^n\}$, the collision detection function $\Pi(\mathcal{S}, B, \mathbf{V})$ is defined as follows, where $B$ represents the bounding box  of an inserted object:

\begin{equation}
\begin{aligned}
\Pi(\mathcal{S}, B, \mathbf{V}) = \begin{cases}
1, & \text{if } \exists i \in \{0, \dots, x\} \text{:} \exists B' \in \mathcal{B}_i \text{:} \\
& \quad \kappa(\text{Move}(B, i \cdot \mathbf{V}), B') = 1 \\
0, & \text{otherwise.} 
\end{cases}
\end{aligned}
\end{equation}

Here, $\text{Move}(B, i \cdot \mathbf{V})$ represents moving the bounding box $B$ along the velocity vector $i\cdot \mathbf{V}$, and $\kappa(B_1, B_2)$ indicates whether bounding boxes $B_1$ and $B_2$ collide. Specifically, two bounding boxes are considered to collide if their projections on the XY plane intersect.

\section{Experiments}

\begin{table}[t]
\tabcolsep=0.18cm
\centering
\caption{Comparisons of 3D tracking task. }
\small
\begin{tabularx}{0.95\linewidth}{l c c c c}
\toprule
Method & {AMOTA $\uparrow$} & AMOTP $\downarrow$ & MOTA $\uparrow$ & Recall $\uparrow$ \\
\midrule
CenterPoint & 65.4 & \textbf{57.3} & 56.2 & 69.0\\
 + D-Aug  & \textbf{66.4} & 57.4  & \textbf{56.4} & \textbf{69.7}\\
\bottomrule
\end{tabularx}
\label{label4}
\end{table}

\subsection{Experimental Settings}
\subsubsection{Dataset.}
The nuScenes \cite{nuScenes} dataset comprises 1,000 driving sequences, with 700, 150, and 150 sequences allocated for training, validation, and testing, respectively. Each sequence spans approximately 20 seconds at 20 frames per second (FPS). The dataset provides calibrated vehicle attitude information and bounding box annotations across 10 classes with long-tailed distributions.



\subsubsection{Metrics.}
For the 3D detection task, mean Average Precision (mAP) and nuScenes Detection Score (NDS) are employed as evaluation metrics. Unlike the conventional 3D Intersection over Union (IoU), mAP is computed based on the aerial view, representing the average of AP across various classes. NDS integrates mAP with additional metrics such as translation, scale, orientation, and velocity into a weighted average. For further details about NDS, please refer to \cite{nuScenes}. In the 3D tracking task, MOTA measures the overall accuracy of multi-object tracking. AMOTA \cite{weng20203d}, which averages MOTA across various IoU thresholds, offers a more comprehensive representation of the tracking capabilities. Meanwhile, AMOTP provides a complementary measure by concentrating on tracking precision. Additionally, the recall metric is considered in 3D tracking experiments.

\subsubsection{Baselines.}
To showcase the effectiveness of our method, we apply our proposed D-Aug to several state-of-the-art (SOTA) methods for 3D object detection and tracking: CBGS \cite{zhu2019class}, PointPillars \cite{pointpillars}, CenterPoint \cite{Yin_Zhou_Krahenbuhl_2021}, VoxelNeXt \cite{Chen_Liu_Zhang_Qi_Jia}, and TransFusion-L \cite{transfusion}. Unfortunately, some of these methods \cite{zhu2019class, pointpillars, Chen_Liu_Zhang_Qi_Jia, transfusion} do not provide code for the 3D tracking task. Therefore, we exclusively conduct experiments related to the 3D tracking task using CenterPoint.


\subsection{Comparisons}

\subsubsection{3D Detection} 
The comparisons for the 3D detection task are presented in Table \ref{label2}. The results highlight performance improvements in most cases when D-Aug is employed. Notably, enhancements are observed across nearly all metrics for CenterPoint \cite{Yin_Zhou_Krahenbuhl_2021}. Significantly, enhancements are observed across all baseline methods, particularly in classes such as "Car", "Tru", "Bus", and "Tra", which share a common characteristic: they typically exhibit higher speeds than other classes. It is evident that dynamic objects derive greater benefits from D-Aug augmentation. Additionally, employing D-Aug with TransFusion-L \cite{transfusion} yields the best performance in both mAP and NDS.

\subsubsection{3D Tracking} 
To further illustrate the efficacy of D-Aug, we conduct comparative experiments on the 3D tracking task. As depicted in Table \ref{label4}, the results indicate performance enhancement, particularly in AMOTA, where D-Aug achieves a 1.0\% increase over the original method. The results solidify the efficacy of D-Aug in the 3D tracking task, as it accounts for the continuity within dynamic scenes.

\subsection{Ablation studies}
\begin{table}[t]
\tabcolsep=0.12cm
\centering
\caption{Ablation studies on the object insertion method}
\small
\begin{tabularx}{0.95\linewidth}{l c c c c}
\toprule
Method & mAP $\uparrow$ & NDS $\uparrow$ & AMOTA $\uparrow$ & AMOTP $\downarrow$ \\
\midrule
Random & 59.39 & 66.94 & 65.9 & 57.2 \\
\hline
Reference-guide & \textbf{59.68} & \textbf{67.16} & \textbf{66.4} & 57.4 \\
- road identification  & 59.63 & 66.89 & 65.8 & \textbf{56.7} \\
\bottomrule
\end{tabularx}
\label{label5}
\end{table}

\begin{table}[t]
\centering
\small
\caption{Ablation studies on the road identification}
\begin{tabularx}{0.95\linewidth}{p{70pt} c c}
\toprule
Method & {Cumtime $\downarrow$} & {Percall $\downarrow$} \\
\midrule
Layer filtering & 1112.04s & 0.881s \\
Pixel-level  & \textbf{39.79s} & \textbf{0.011s} \\
\bottomrule
\multicolumn{3}{p{230pt}}{The 'Cumtime' indicates the total time spent within a data augmentation process, while 'Percall' indicates the average time spent per call.}
\end{tabularx}
\label{tab: comparison}
\end{table}

To assess the efficacy of each proposed component, we perform ablation studies on both 3D object detection and tracking tasks. Our ablation studies are conducted exclusively on CenterPoint since it provides code for the 3D tracking task.

\subsubsection{Object Insertion Method}
In object insertion, we propose a reference-guided insertion algorithm instead of randomly choosing an insertion position within the scene. Additionally, pixel-level road identification is introduced to ensure the validity of the insertion position. We validate our proposed insertion method through ablation studies, as shown in Table \ref{label5}. We also conduct an efficiency comparison between pixel-level road identification and the layer filtering method offered in nuScenesMap-API, with the results shown in Table \ref{tab: comparison}.


\begin{table}[t]
    \centering
    \tabcolsep=0.14cm
    \small 
    \caption{Ablation studies on the voxel size}
    \begin{tabularx}{0.95\linewidth}{l c c c c c}
        \toprule
        \multirow{1}{*}{Size}  & Method & mAP $\uparrow$ & NDS $\uparrow$ & AMOTA $\uparrow$ & AMOTP $\downarrow$ \\
        \midrule
        \multirow{2}{*}{0.075}  & CenterPoint & 59.10 & 66.69 & 65.4 & \textbf{57.3}\\
        & +D-Aug & \textbf{59.68} & \textbf{67.16} & \textbf{66.4} & 57.4 \\
        \hline
        \multirow{2}{*}{0.100}  & CenterPoint& 55.15 & 64.16 & 61.1 & 64.4 \\
        & +D-Aug& \textbf{56.15} & \textbf{64.61} & \textbf{62.6} & \textbf{61.0} \\
        \hline
        \multirow{2}{*}{0.200}  & CenterPoint & 51.68 & 60.61 & 58.6 & \textbf{65.4} \\
         & +D-Aug & \textbf{52.16} & \textbf{61.05} & \textbf{59.2} & 65.6 \\
        \bottomrule
    \end{tabularx}
    \label{tab:multiro}
\end{table}

\subsubsection{Voxel Size}
To efficiently manage substantial volumes of point cloud data, conversion into voxels is essential, as they offer a discrete, grid-based representation. While smaller voxels preserve finer details, they demand increased computational resources. To underscore the versatility of D-Aug, we conduct experiments with varying voxel sizes, as depicted in Table~\ref{tab:multiro}. Encouragingly, the results demonstrate consistent enhancements across different voxel sizes, affirming the generalizability of our approach.

\subsubsection{Quantity of Inserted Objects}
The quantity of inserted objects can significantly impact performance. Therefore, we vary the number of inserted objects from 1 to 8 and analyze the results. As depicted in Table \ref{label8}, we find that 5 objects yield optimal performance. This finding suggests that an excessive number of inserted objects may introduce unrealistic elements, while too few objects might not fully exploit the benefits of data augmentation. 

\begin{table}[t]
\tabcolsep=0.35cm
\centering
\caption{Ablation Studies on the quantity of inserted objects}
\small
\begin{tabularx}{0.95\linewidth}{c c c c c}
\toprule
Num & mAP $\uparrow$ & NDS $\uparrow$ & AMOTA $\uparrow$ & AMOTP $\downarrow$ \\
\midrule
1 & 59.46 & 66.58 & 65.7 & 58.5 \\
3  & 59.36 & 66.75 & 60.4 & 62.2 \\
5  & \textbf{59.68} & \textbf{67.16} & \textbf{66.4} & \textbf{57.4} \\
8  & 59.38 & 66.92 & 65.3 & 58.4 \\
\bottomrule
\end{tabularx}
\label{label8}
\end{table}
\section{Conclusion}
This paper introduces D-Aug, a specialized LiDAR data augmentation method designed for dynamic scenes. D-Aug entails extracting objects from LiDAR data and seamlessly inserting them into dynamic scenes while preserving their coherence across consecutive frames. Validating insertion positions is ensured through precise pixel-level road identification and reference-guided insertion strategy. Our experiments validate the effectiveness of D-Aug in enhancing performance across nuScenes detection and tracking benchmarks. Nonetheless, occlusion within the point cloud remains a challenge, suggesting that addressing post-insertion occlusion represents a promising direction for future work. 


\bibliographystyle{ieeetr}
\balance
\bibliography{main}
 
\end{document}